\documentclass[letterpaper]{article} 
\usepackage{aaai25}  
\usepackage{times}  
\usepackage{helvet}  
\usepackage{courier}  
\usepackage[hyphens]{url}  
\usepackage{graphicx} 
\usepackage{natbib}  
\usepackage{caption} 
\usepackage{algorithm}
\usepackage{algorithmic}

\usepackage{newfloat}
\usepackage{listings}
\DeclareCaptionStyle{ruled}{labelfont=normalfont,labelsep=colon,strut=off} 
\floatstyle{ruled}
\newfloat{listing}{tb}{lst}{}
\floatname{listing}{Listing}
\usepackage[american]{babel}
\usepackage{overpic}
\usepackage{amssymb}
\usepackage{amsmath}
\usepackage{multirow}
\usepackage{booktabs}
\usepackage{xcolor}
\usepackage{caption}
\usepackage{subfloat}
\usepackage{bbding}
\usepackage{multirow}
\usepackage{amsmath,amssymb}
\RequirePackage{xspace}
\makeatletter
\DeclareRobustCommand\onedot{\futurelet\@let@token\@onedot}
\def\@onedot{\ifx\@let@token.\else.\null\fi\xspace}

\def\eg{\emph{e.g}\onedot} 
\def\ie{\emph{i.e}\onedot}

\makeatother

\usepackage[capitalize]{cleveref}
\crefname{section}{Sec.}{Secs.}
\Crefname{section}{Section}{Sections}
\Crefname{table}{Table}{Tables}
\crefname{table}{Tab.}{Tabs.}
\Crefname{equation}{Equation}{Equations}
\crefname{equation}{Eqn.}{Eqns.}

\usepackage{bm}

\newcommand{\mat}[1]{\bm{\mathit{#1}}}
\newcommand{\matx}{\mat{x}}
\newcommand{\ii}[1]{\mathit{#1}}

\title{Learning Spatially Decoupled Color Representations for\\Facial Image Colorization}

\author {
    Hangyan Zhu\textsuperscript{\rm 1}, Ming Liu\textsuperscript{\rm 1,*}, Chao Zhou\textsuperscript{\rm 2}, Zifei Yan\textsuperscript{\rm 1}, Kuanquan Wang\textsuperscript{\rm 1}, Wangmeng Zuo\textsuperscript{\rm 1}
}
\affiliations {
    \textsuperscript{\rm 1}Harbin Institute of Technology, Harbin, China,
    \textsuperscript{\rm 2}Shanghai Transsion Co, Ltd., Shanghai, China\\
    cshyzhu@outlook.com, csmliu@oulook.com, wmzuo@hit.edu.cn
}

\usepackage{bibentry}

\begin{document}

\maketitle

\begin{abstract}
	Image colorization methods have shown prominent performance on natural images. However, since humans are more sensitive to faces, existing methods are insufficient to meet the demands when applied to facial images, typically showing unnatural and uneven colorization results.
	In this paper, we investigate the facial image colorization task and find that the problems with facial images can be attributed to an insufficient understanding of facial components.
	As a remedy, by introducing facial component priors, we present a novel facial image colorization framework dubbed FCNet.
	Specifically, we learn a decoupled color representation for each face component (\eg, lips, skin, eyes, and hair) under the guidance of face parsing maps.
	A chromatic and spatial augmentation strategy is presented to facilitate the learning procedure, which requires only grayscale and color facial image pairs.
	After training, the presented FCNet can be naturally applied to facial image colorization with single or multiple reference images.
	To expand the application paradigms to scenarios with no reference images, we further train two alternative modules, which predict the color representations from the grayscale input or a random seed, respectively.
	Extensive experiments show that our method can perform favorably against existing methods in various application scenarios (\ie, no-, single-, and multi-reference facial image colorization).
	The source code and pre-trained models will be publicly available.
\end{abstract}

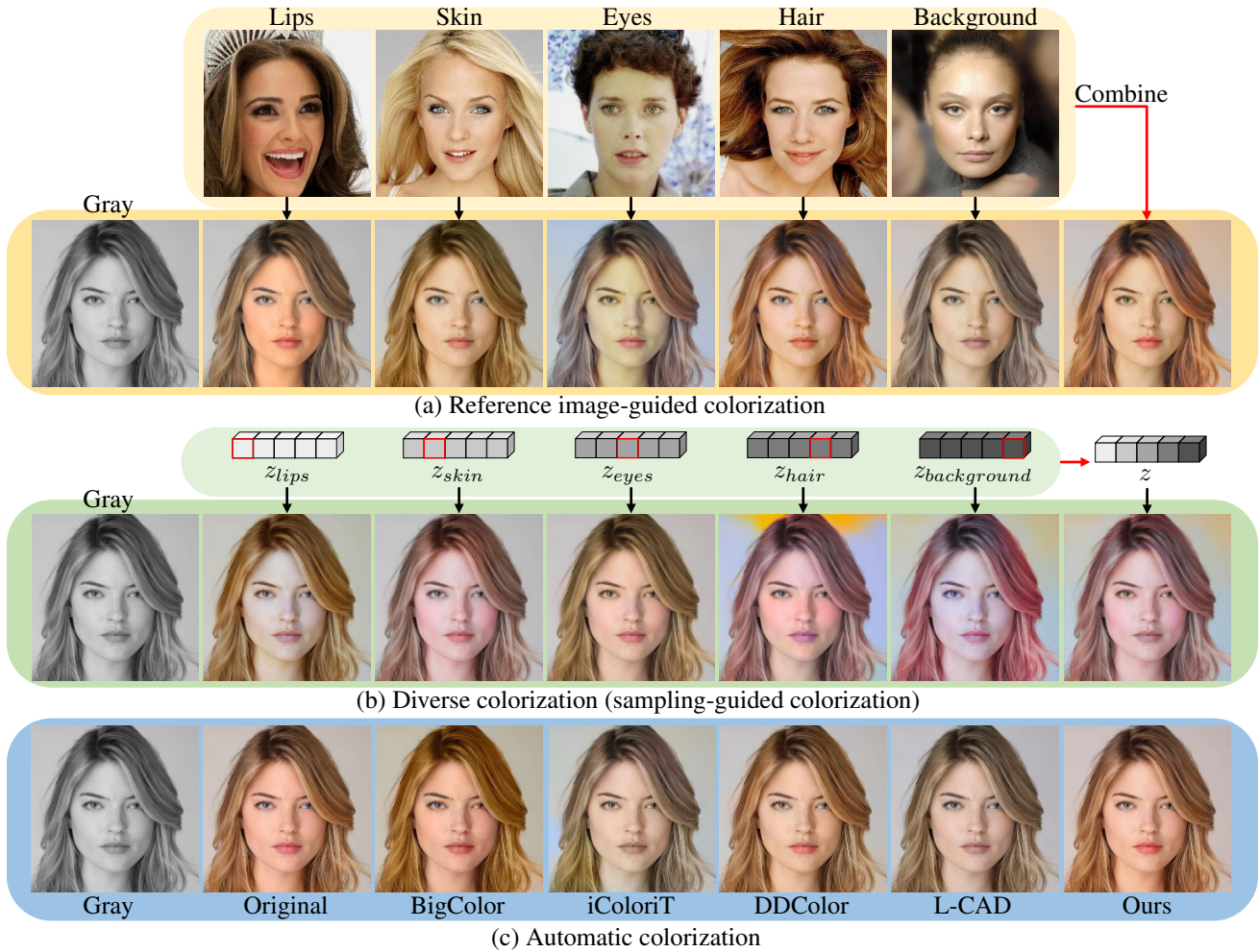
\begin{figure*}[!t]
	\centering
	\begin{overpic}[scale=0.34]{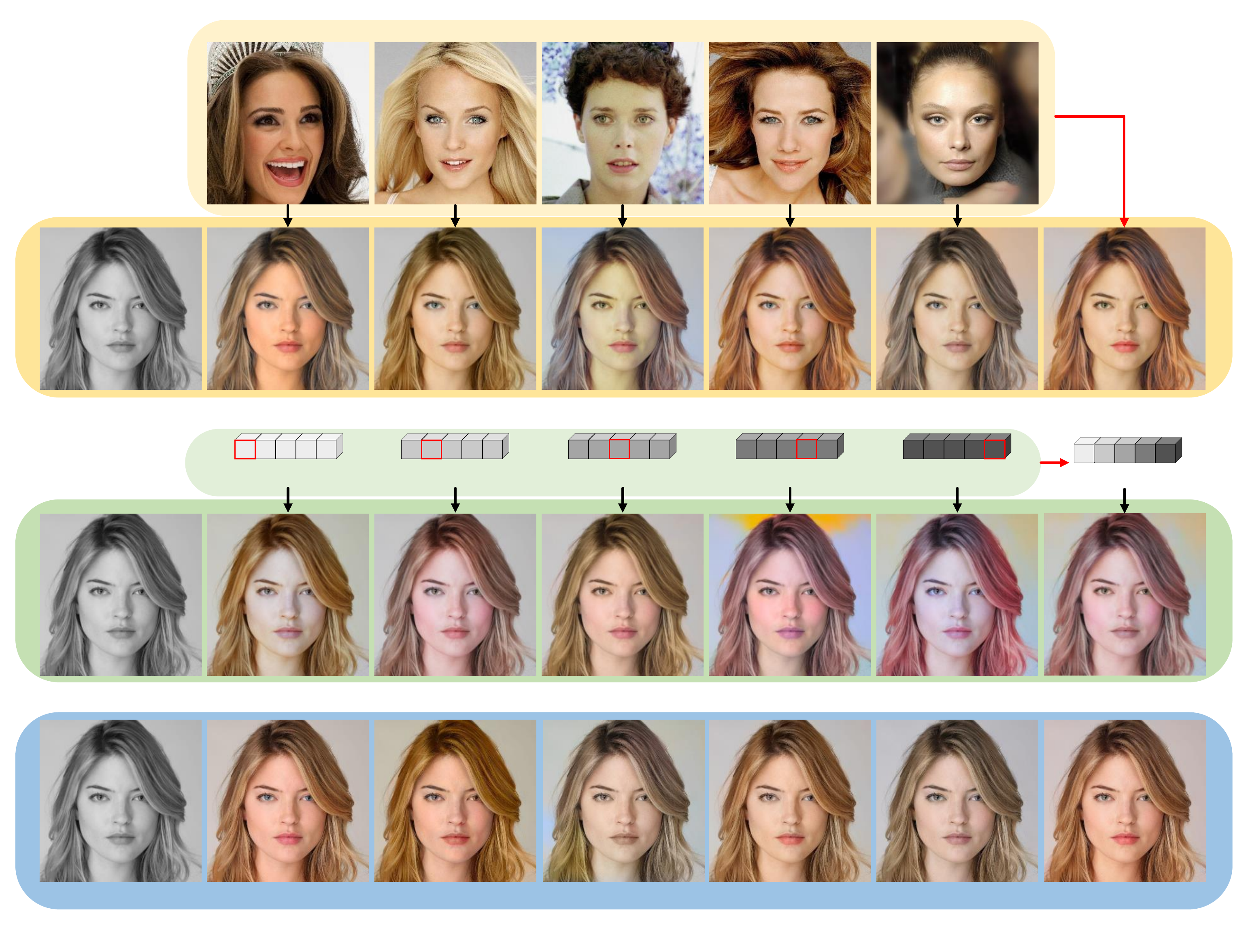} 
		\put(33,43) {(a) Reference image-guided colorization}
		\put(28.5,20) {(b) Diverse colorization (sampling-guided colorization)}
		\put(39,1.5) {(c) Automatic colorization}
		\put(84.5,67.3) {Combine}
		\put(21.7,73.3) {Lips}
		\put(34.7,73.3) {Skin}
		\put(47.7,73.3) {Eyes}
		\put(61.4,73.3) {Hair}
		\put(71.9,73.3) {Background}
		\put(7.2,58.7) {Gray}
		\put(7.2,35.7) {Gray}
		\put(7.2,4) {Gray}
		\put(19.7,4) {Original}
		\put(32.7,4) {BigColor}
		\put(46.5,4) {iColoriT}
		\put(59.5,4) {DDColor}
		\put(73.5,4) {L-CAD}
		\put(21.3,38) {$z_{lips}$}
		\put(34.3,38) {$z_{skin}$}
		\put(47.6,38) {$z_{eyes}$}
		\put(61.0,38) {$z_{hair}$}
		\put(71.9,38) {$z_{background}$}
		\put(89.5,37.8) {$z$}
		\put(88.2,4) {Ours}
	\end{overpic}
	\caption{Our method encompasses three colorization approaches, \ie, single- or multi-reference image-guided colorization in (a), sampling-guided colorization in (b), and automatic colorization in (c). In (a), the first row shows five reference images from different identities, while the second row provides the grayscale input, five colorized images referring to the five reference images, and a result whose colorization for different facial components relies on different reference images. In (b), the results are also generated according to the sampled single or multiple color representations. In (c), we give our results under automatic settings and the results of competing methods.}
	\label{fig:overview_intro}
\end{figure*}

\section{Introduction}
\label{sec:intro}

Image colorization aims at predicting the color for grayscale images, which can be utilized for a variety of tasks such as old photo restoration, advertising, and art creation~\cite{iizuka2019deepremaster,qu2006manga,tsaftaris2014novel,vitoria2020chromagan}.
With years of development, a series of image colorization methods~\cite{ColorfulColorization, deshpande2017learning, GCP, DDcolor, LCAD} have been presented, and user assistance, as well as reference images, are also introduced for user-intended colorization results~\cite{levin2004colorization, chia2011semantic, unicolor, PDNLA-Net}.
These methods have achieved prominent image colorization performance on natural images.

Among various image contents, faces are a category of special objects that should be carefully considered.
On the one hand, faces are very common in images, especially in old photos where faces are the main subject of photography in the early years, which poses a great demand for facial image colorization.
On the other hand, faces are one of the most familiar objects for us. Thus, humans are sensitive to faces and can perceive subtle anomalies in facial images, making facial image colorization quite challenging.
In our experiments, when applying existing image colorization methods to faces, they tend to produce unnatural and uneven results, which are insufficient to meet the user requirements.

The failure of existing methods on facial images motivates us to develop a face-oriented image colorization method.
In this paper, we investigate the facial image colorization task and find that the problem of existing methods on facial images mainly lies in an insufficient understanding of facial components.
For example, sometimes the generated colors of the upper and lower parts of the cheek are significantly different, which can be attributed to the image colorization models not realizing that the cheek is a whole component and the color is usually consistent.
To remedy such a phenomenon and generate visually pleasing results, we propose a novel facial image colorization framework dubbed FCNet by introducing facial component priors.

Specifically, we force the model to possess the knowledge of face structures and learn a decoupled color representation for each facial component.
To begin with, a pre-trained face parsing model is introduced, which provides the segmentation of facial components as a prior.
The explicitly introduced facial component prior greatly alleviates the burden of the colorization modules.
Then, an encoder can be deployed to learn the color representations.
However, even with the face component priors, the color representations of different face components are still entangled, which may cause the problem of color infiltration (\ie, the color of a component is affected by another one).

For learning a decoupled color representation for each facial component, we present a data augmentation strategy.
Specifically, the color image is augmented to different versions with chromatic and spatial transforms, and color representations are extracted from different augmented versions via a shared encoder.
By slicing and reorganizing these color representations as a new one, the color representation from each version can be restricted to contribute only to a particular facial component of the colorization result.
The supervision is also a composite color image combined from different augmented versions.

After training, the model can be naturally utilized for reference-based facial image colorization, under both single- and multi-reference settings.
Considering that the references are not always easy to prepare, we further expand the application paradigm of the presented FCNet to scenarios with no reference images.
For automatic colorization, another encoder is trained to predict the color representation from the input grayscale image (and the face parsing maps).
To further introduce randomness and diversity into the prediction results, we also model the color distribution of each facial component via normalizing flow models~\cite{NICE,RealNVP,GLOW}, which can randomly sample a color representation from the color distribution during the inference stage. The exemplary results of the three colorization approaches are illustrated in \cref{fig:overview_intro}.

For evaluating the presented FCNet, comprehensive experiments on facial image colorization are conducted under different settings (\ie, no-, single-, and multi-reference).
Both quantitative and qualitative results show the effectiveness of our FCNet.
To sum up, the contributions of this paper are as follows.
\begin{itemize} 
	\item Our proposed FCNet decouples the colors of different facial components with the help of facial component priors and a delicately designed data augmentation strategy, enabling individual color control for each facial component and ensuring the controllability of the network.
	\item Multiple application paradigms, including automatic colorization, diverse colorization, and single- and multi-reference-based colorization, are implemented, providing a flexible facial image colorization solution.
	\item Experiments show that our method not only achieves precise control over the colors of different facial components but also produces colorization results that are more vibrant and diverse than previous approaches. 
\end{itemize}

\section{Related Work}
\label{sec:related_works}
\paragraph{Early user-assisted colorization.}
Early colorization methods were often user-assisted. One such approach was scribble-based methods, which involved combining segmentation algorithms with manual color assignment by users for each region of the image, resembling a coloring book-like process. Starting from the~\cite{levin2004colorization}, scribble-based colorization was optimized based on spatial continuity using least squares optimization. Subsequent methods~\cite{ironi2005colorization,liu2008intrinsic} further advanced this technique by introducing improvements and refinements based on the initial framework. Although these approaches can produce visually pleasing results through meticulous user interactions, they often require substantial user efforts. To address this limitation, \cite{zhang2017real} addressed this limitation by using sparse color points and a neural network for colorization. Other approaches explored using global hints such as color palettes~\cite{bahng2018coloring, chang2015palette} as constraints instead of dense color points.

\paragraph{Reference-based methods.}Early approaches in this category transferred the color statistics from a reference image to a grayscale image by utilizing spatial consistency~\cite{chia2011semantic,gupta2012image}, low-level similarity measures~\cite{liu2008intrinsic}, or semantic features~\cite{charpiat2008automatic}. In recent years, notable advancements have been made by combining deep learning techniques, as demonstrated by prominent works such as~\cite{he2018deep,xu2020stylization,wct2,Gray2colornet,TFcolor,ke2023neural,PDNLA-Net}. Although this method can achieve visually appealing colorization results and provides control over the image colors, it falls short in precisely specifying the colors of individual objects within each image. Additionally, finding suitable reference images can be challenging.

\paragraph{Automatic colorization.}Since the advent of~\cite{cheng2015deep}, data-driven deep learning methods for colorization have gained significant traction. Notably,~\cite{ColorfulColorization} introduced the rebalanced multinomial cross-entropy approach to effectively capture the color distribution of individual pixels, while~\cite{deshpande2017learning} incorporated a variational autoencoder (VAE) structure to enhance colorization diversity. In recent years,~\cite{Instcolor} integrated detection models to leverage detection boxes as informative priors, while~\cite{ zhao2018pixel,zhao2020pixelated} employed segmentation masks as pixel-level priors. Moreover,~\cite{BigColor,GCP}  harnessed pre-trained generative adversarial networks (GANs) to provide valuable color priors, and~\cite{kumar2021colorization,ji2022colorformer,ct2,DDcolor} explored the integration of the Vision Transformer (ViT)~\cite{dosovitskiy2020image} architecture to offer color priors.
Recent work~\cite{LCAD,liang2024control,zhang2023controlvideo} has proposed a novel automatic colorization methodology leveraging pre-trained diffusion-based models to provide realistic color priors.

\begin{figure*}[tb]
	\centering
	\begin{overpic}[scale=0.36]{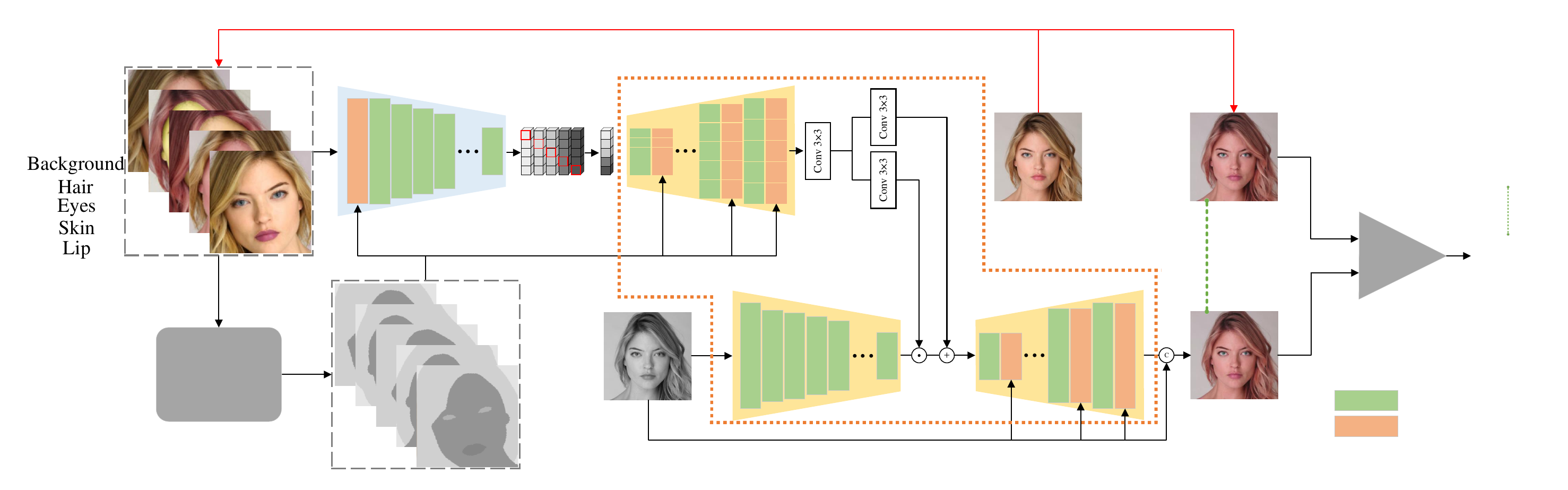}  
		\put(2.0,25.0) {$\{\matx_\ii{ref}^i\}$}
		\put(35.8,27.7) {Chromatic and Spatial Augmentation}
		\put(68.0,27.7) {Chromatic}
		\put(66.8,25.7) {Augmentation}
		\put(26,25.7) {$g$}
		\put(40,24.7) {$f$}
		\put(43,16.8) {$G_w$}
		\put(89.5,5.2) {Resblock}
		\put(89.5,3.4) {SPADE}
		\put(36.6,8.1) {$\mat{x}^l$}
		\put(32.9,18.1) {$\{\mat{w}^i\}$}
		\put(37.75,19.1) {$\mat{w}$}
		\put(12.1,8.8) {Face}
		\put(11.6,6.8) {Parser}
		\put(13.1,4.8) {$\mathcal{P}$}
		\put(50.1,12.3) {$E_{Gray}$}
		\put(65.3,12.1) {$G$}
		\put(64.8,14.5) {$\mathcal{L} _{1}, \mathcal{L} _{perc}, \mathcal{L} _{cyc}$}
		\put(87.8,14.3) {$\mathcal{D}$}
		\put(78,16.9) {$\tilde{\matx}$}
		\put(65.5,16.9) {$\matx$}
		\put(78,4.2) {$\hat{\matx}$}
		\put(81.6,21.5) {True}
		\put(81.6,7.0) {False}
		\put(94.3,15.0) {True/}
		\put(94.3,13.3) {False}
		\put(94.4,20.0) {$\mathcal{L} _{adv}$}
	\end{overpic}
	\caption{Overview and main training phase of our method. As depicted in the figure, the red arrows represent the data augmentation process. The orange dashed enclosure highlights the colorization network $f$. $\mathcal{D}$ is the discriminator.}
	\label{fig:overview}
\end{figure*}
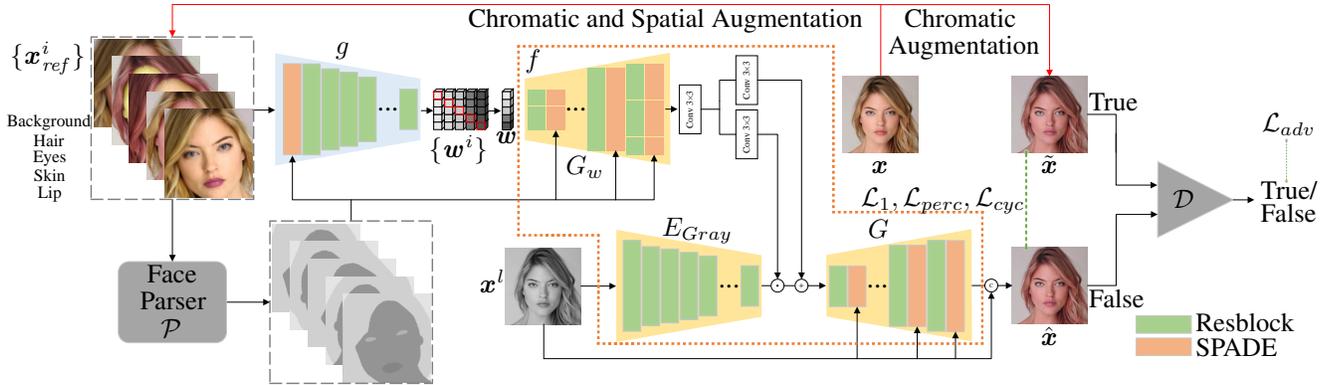

\section{Method}
\label{sec:method}

\subsection{Overview}
In our observation, the main problem with existing image colorization methods on facial images is typically manifested in unnatural and uneven colorization results, which is mainly attributed to the inadequate understanding of facial parts by existing methods.
Therefore, the primary objective of this work is to introduce facial component priors into the image colorization procedure.
To achieve such an objective, we develop an intuitive solution that simply deploys a face parsing model, which can explicitly provide information about the regions of the facial components.
Furthermore, a chromatic and spatial augmentation strategy is presented to force the model to align with the facial component regions.
Considering the color consistency of the facial components, we choose to represent the color of each facial component via a low-dimension latent representation, which is then expanded according to the input face structure.

Based on the above analyses, our model is designed with two collaborative modules, \ie, the color representation branch and the colorization network.
As shown in \cref{fig:overview}, given a grayscale image $\mat{x}^l\in\mathbb{R}^{H \times W \times 1}$, the color representation branch extracts a color representation from several reference images (each reference image $\mat{x}^{ab}\in\mathbb{R}^{H \times W \times 2}$ has only the ab channels of the CIE-Lab color space).
Then, the colorization network can generate the final output given the grayscale input and the color representation.
This framework enables each module to focus only on specific problems, avoiding the interference between color prediction and colorization, laying the foundation for expanding the application scenarios of the model.
In the following, we will provide more details about the network.

\subsection{Color Representation Branch}
\paragraph{Facial Component Priors.}
The main purpose of the color representation branch is to indicate the appropriate color for each facial component.
Formally speaking, given a reference image $\mat{x}_\ii{ref}$, we hope that the color representation branch can capture the color of facial components in $\mat{x}_\ii{ref}$, resulting in a group of color representations $\mat{w}_\ii{c}$, where $c\in\{\ii{lips}, \ii{skin}, \ii{eyes}, \ii{hair}, \ii{background}\}$.
In order to equip the model with facial component priors, a face parser $\mathcal{P}$ is introduced, which provides the masks of the facial components (\ie, $\mat{m}_\ii{ref}=\mathcal{P}(x_\ii{ref})$).

Considering the colorization pipeline, the final result $\hat{\mat{x}}=f(\mat{x}^l, g(\mat{x}_\ii{ref}, \mat{m}_\ii{ref}))$, where $f$ and $g$ denote the colorization network and the color representation branch, respectively.
There are no constraints on the decoupling of the color representations for these facial components.
In other words, even with the facial component priors (\ie, $\mat{m}_\ii{ref}$), the color representations of different facial components are still entangled.

\paragraph{Data Augmentation.}
To learn a decoupled color representation, we present a data augmentation strategy based on existing training data (where only a color image $\mat{x}$ and its grayscale version $\mat{x}^l$ are available).
Specifically, as shown in \cref{fig:overview}, the color image $\matx$ is augmented from both chromatic and spatial perspectives, resulting in five augmented images $\{\matx_\ii{ref}^i\}_{i=0}^4$, and we take a certain facial component from each $\matx_\ii{ref}^i$ to make up a corresponding ground truth $\tilde{\matx}$.
In this way, the colorization representation branch can extract a group of colorization representations $\{\mat{w}^i\}$ from these augmented images, and each $\mat{w}^i$ provide only a certain component in the final $\mat{w}$, \ie, $\mat{w}=[\mat{w}^0_\ii{lips},\mat{w}^1_\ii{skin},\mat{w}^2_\ii{eyes},\mat{w}^3_\ii{hair},\mat{w}^4_\ii{background}]$.

Without the chromatic augmentation, the spatial augmentation becomes meaningless, where the model can simply rely on a certain reference image.
Without the spatial augmentation, the reference images with different postures or those from other identities are unable to be utilized in the inference stage.
With both the chromatic and spatial augmentation, the color representation branch can be constrained to obtain the decoupled color representation $\mat{w}$.

\subsection{Colorization Network}
Given the color representation $\mat{w}$, the colorization network $f$ can perform the colorization procedure.
As shown in \cref{fig:overview}, $f$ consists of two decoders and one encoder.
The color representation decoder, denoted as $G_w$, processes the color representation $\mat{w}$ obtained from the color representation branch, and the resultant feature $F_w$ serves as a guidance of the colorization. 
Notably, $G_w$ employs grouped design (\ie, group convolution) to process the respective parts of $\mat{w}$ individually.
The gray image encoder, denoted as $E_\ii{Gray}$, extracts the feature $F_{l}$ from the input grayscale image.
The grayscale feature $F_l$ is modulated by the color representation feature $F_w$ via an affine transform (\ie, the multiplication and addition operations).
Finally, the decoder $G$ generates the ab channel of the final colorization result $\hat{\matx}$, where the grayscale input participates in the decoding process and serves as the l channel of $\hat{\matx}$.
More details about the model can be found in the supplementary materials.

\begin{figure*}[tb] \centering
	
	\begin{overpic}[scale=0.41]{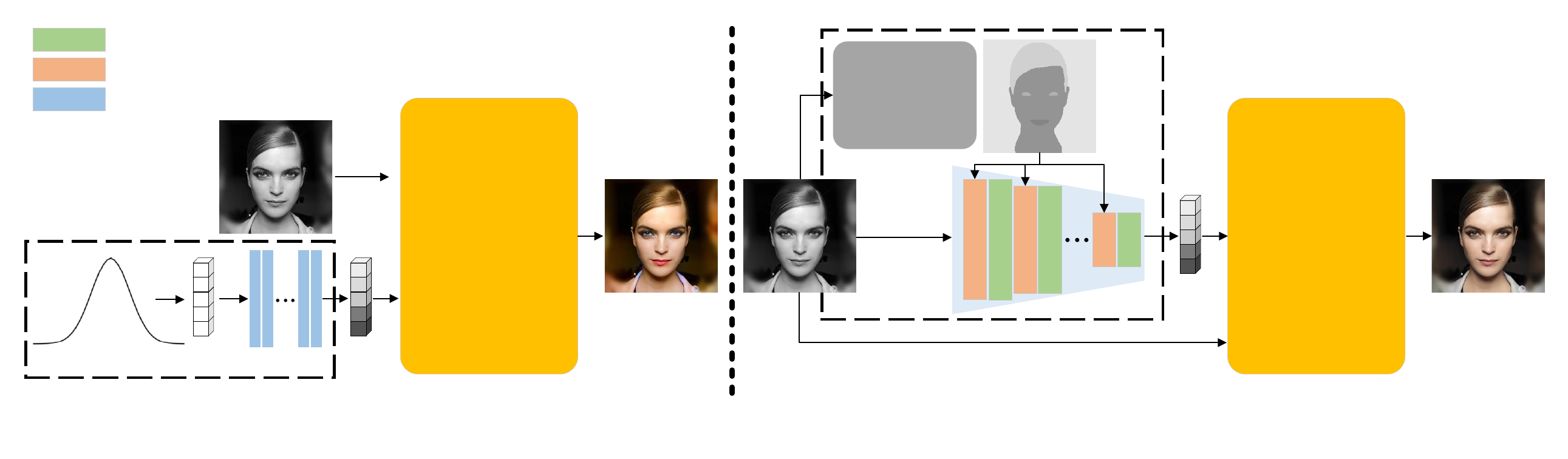} 
		\put(7.2,26.3) {Resblock}
		\put(7.2,24.3) {SPADE}
		\put(7.2,22.4) {Linear flow block}
		\put(53.2,23.4) {Face Parser}
		\put(57.0,21.4) {$\mathcal{P}$}
		\put(70.0,22.4) {$\mat{m}^l$}
		\put(13.5,2) {(a) Diverse Colorization}
		\put(61.5,2) {(b) Automatic Colorization}
		\put(16.3,6) {$g_{flow}$}
		\put(22.1,6.7) {$\mat{w}$}
		\put(12,17.5) {$\mat{x}^{l}$}
		\put(41.5,9.0) {$\mat{\hat{x}}$}
		\put(48.7,9.0) {$\mat{x}^{l}$}
		\put(69.2,9.7) {$g_\ii{auto}$}
		\put(75.1,10.8) {$\mat{w}$}
		\put(94.3,9.0) {$\mat{\hat{x}}$}
		\put(26.3,15.5) {Colorization}
		\put(27.8,13.5) {Network}
		\put(30.8,11.5) {$f$}
		\put(79.0,15.5) {Colorization}
		\put(80.5,13.5) {Network}
		\put(83.5,11.5) {$f$}
	\end{overpic}
	\caption{Two application paradigms for No-reference scenarios. (a) denotes the Diverse Colorization, wherein the $g_\ii{flow}$ can randomly generate a color representation for a given grayscale image $\mat{x}^{l}$ by sampling from a known probability distribution. (b) denotes the Automatic Colorization, wherein the $g_\ii{auto}$ encodes $\mat{x}^{l}$ and its associated face parsing map $\mat{m}^l$ to generate the color representation $\mat{w}$.}\label{fig:train_phase}
\end{figure*}
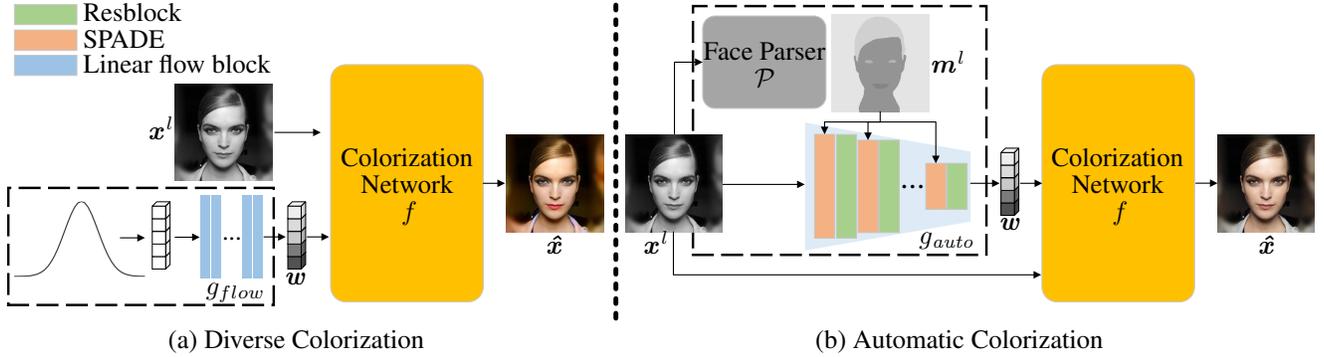

\subsection{Learning Objectives}
With the color representation branch and the colorization network, the model can be trained simply following existing colorization methods, where $\ell_1$ loss, perceptual loss~\cite{johnson2016perceptual}, cycle-consistency loss~\cite{zhu2017unpaired}, and adversarial loss~\cite{heusel2017gans} are utilized.
The only thing worth noting is that, we take all possible data for training, including (1) the original color image $\matx$, (2) each augmented color image $\matx_\ii{ref}^i$, and (3) the composite color image $\tilde{\matx}$.
Therefore, the learning objective can be defined by,
\begin{equation}
	\mathcal{L}=\mathcal{L}_\ii{adv}+\alpha\mathcal{L}_1+\beta\mathcal{L}_\ii{perc}+\gamma\mathcal{L}_\ii{cyc},
\end{equation}
where $\alpha$, $\beta$, and $\gamma$ are hyper-parameters that are empirically set to 1.0, 0.05, and 1.0, respectively.
More details about the loss functions can be found in the supplementary materials.

\subsection{Extension to No-reference Scenarios}
Given the above network design and learning objectives, the model can be trained for both single- and multi-reference facial image colorization tasks.
However, references are not always easy to prepare, which motivates us to extend the model to no-reference scenarios.
Thanks to the design of deploying a separate color representation branch, we can safely replace the color representation branch with the alternative modules, which implements the following two application paradigms.

\paragraph{Automatic Colorization.}
Once the FCNet is trained, the color representation $\mat{w}$ becomes meaningful. Therefore, as shown in \cref{fig:train_phase}, we can directly learn $\mat{w}$ with another encoder $g_\ii{auto}$, which predicts a $\mat{w}$ from the grayscale input $\matx^l$ and the face parsing map $\mat{m}^l$.

\paragraph{Diverse Colorization.}
Given the meaningful $\mat{w}$, we can also learn the distribution of the color representation space with a normalizing flow network $g_\ii{flow}$.
In this way, $g_\ii{flow}$ can randomly generate a color representation for a given grayscale input $\matx^l$, which ensures the diversity and randomness.
The network structure and learning objective details of $g_\ii{auto}$ and $g_\ii{flow}$ can be found in the supplementary materials.

\section{Experiments}
\label{sec:Experiments}

\subsection{Experimental Setting}
\paragraph{Datasets.} We conduct our experiments on two aligned facial image datasets. The Flickr-Faces-HQ (FFHQ) dataset~\cite{karras2019style} comprises over 70,000 well-aligned facial images of diverse individuals, which we employ as the training set. Conversely, the CelebA-HQ dataset~\cite{karras2017progressive} consists of 30,000 high-resolution aligned facial images of celebrity subjects, which we leverage as the test set.

\paragraph{Evaluation metrics.} Consistent with the previous colorization works, we primarily utilize the Fréchet inception distance (FID)~\cite{heusel2017gans} and colorfulness score (CF) as evaluation metrics to assess the effectiveness of our proposed method. For supplementary reference, we also report the Peak Signal-to-Noise Ratio (PSNR) \cite{huynh2008scope} and Structural Similarity Index Measure (SSIM) \cite{wang2004image} to comprehensively assess the visual perception quality and detailed characteristics of the colorization results.

\paragraph{Implementation details.} We train our network with the Adam optimizer~\cite{kingma2014adam}, and set the hyperparameters $\beta_1=0.5$ and $\beta_2=0.999$. During the training of our colorization pipeline, we initialize the learning rate to $5\times10^{-5}$ and utilize a batch size of $4$. Conversely, when training the subcomponents $g_{flow}$ and $g_{auto}$, we initialize the learning rate to $1\times10^{-3}$ and employ a batch size of $16$. Throughout the training process, we resize the training images to dimensions of $256\times256$ pixels and execute the training procedure for a total of 50 epochs over the entirety of the training set. All experiments are conducted on a single NVIDIA RTX A6000 GPU. For detailed information regarding the experimental hardware and software environment, as well as the implementation details, please refer to the supplementary materials.

\begin{figure*}[tb] \centering
	
	\begin{overpic}[scale=0.325]{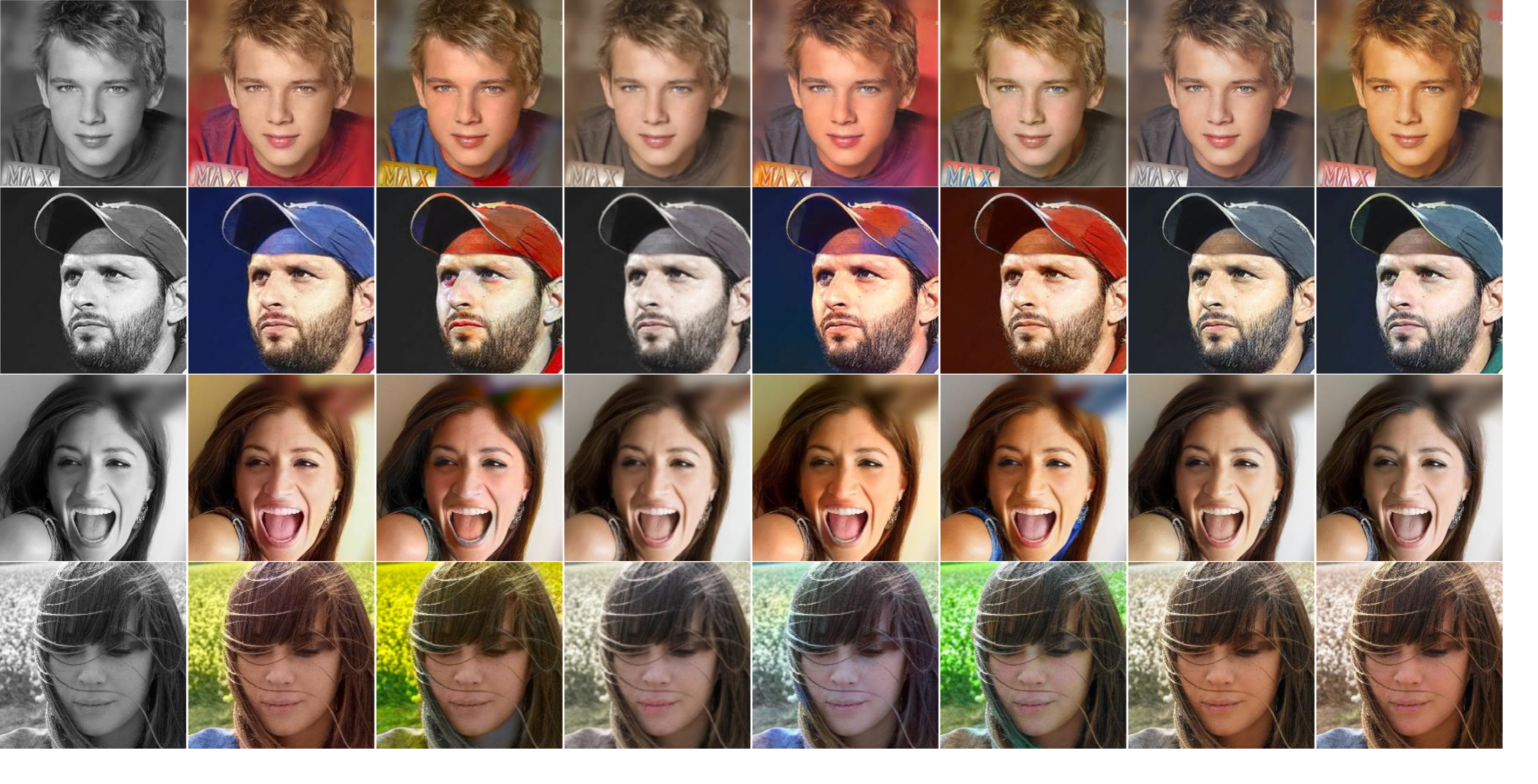}
		\put(4,0.6) {Gray}
		\put(17,0.6) {CT2}
		\put(27.5,0.6) {BigColor}
		\put(40.3,0.6) {iColoriT}
		\put(52.5,0.6) {DDColor}
		\put(65.5,0.6) {L-CAD}
		\put(79.1,0.6) {Ours}
		\put(90.3,0.6) {Original}
	\end{overpic}
	\caption{The qualitative results of our method and automatic colorization baseline methods.} \label{fig:Qualitative}
\end{figure*}

\begin{figure*}[tb] \centering
	\begin{overpic}[scale=0.29]{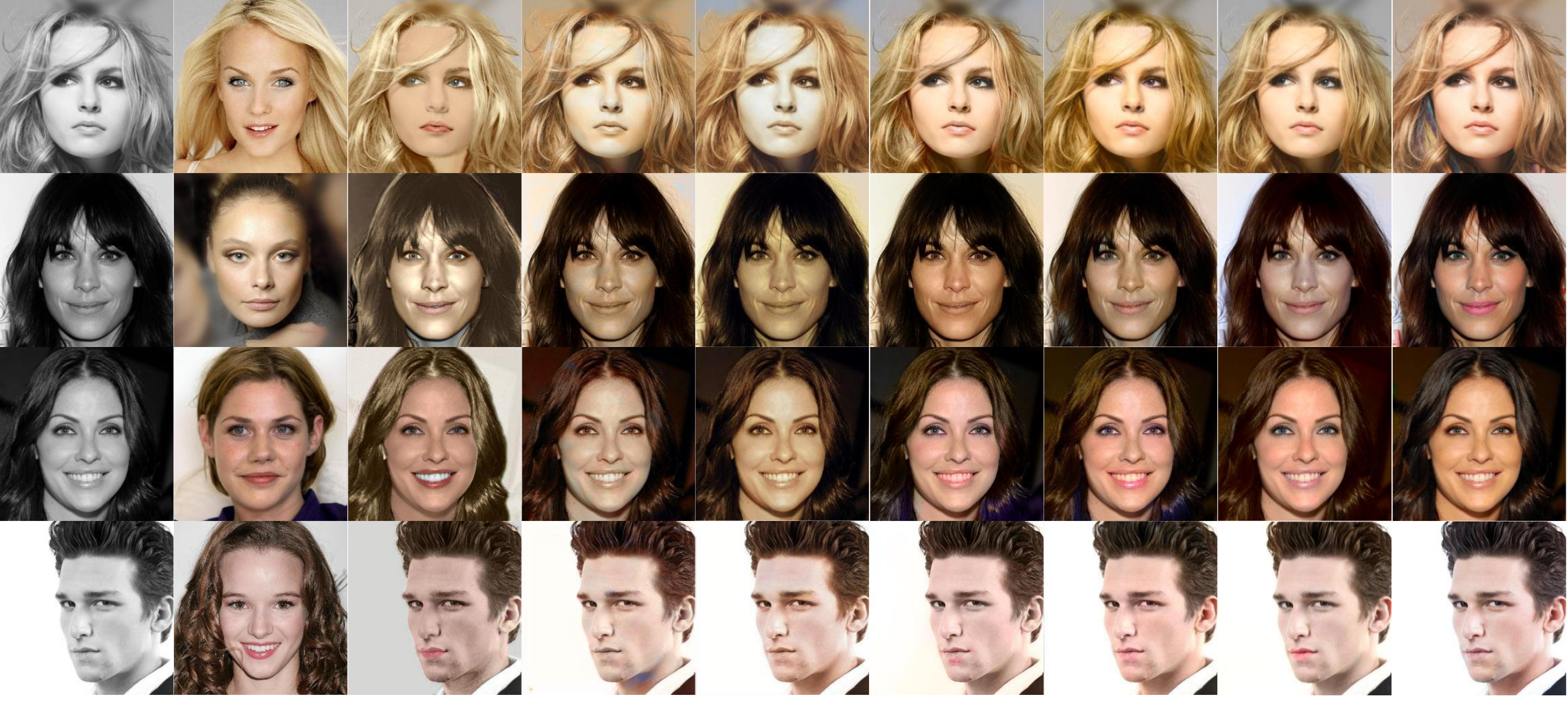} 
		\put(3.5,0.5) {Gray}
		\put(12.7,0.5) {Reference}
		\put(24.8,0.5) {WCT2}
		\put(32.8,0.5) {Gray2ColorNet}
		\put(46.8,0.5) {TFColor}
		\put(57.5,0.5) {Unicolor}
		\put(67.1,0.5) {PDNLA-Net}
		\put(81.2,0.5) {Ours}
		\put(91.1,0.5) {Original}
	\end{overpic}
	\caption{The qualitative results of our method and reference image-based colorization baseline methods.
	} \label{fig:Qualitative_ref}
\end{figure*}
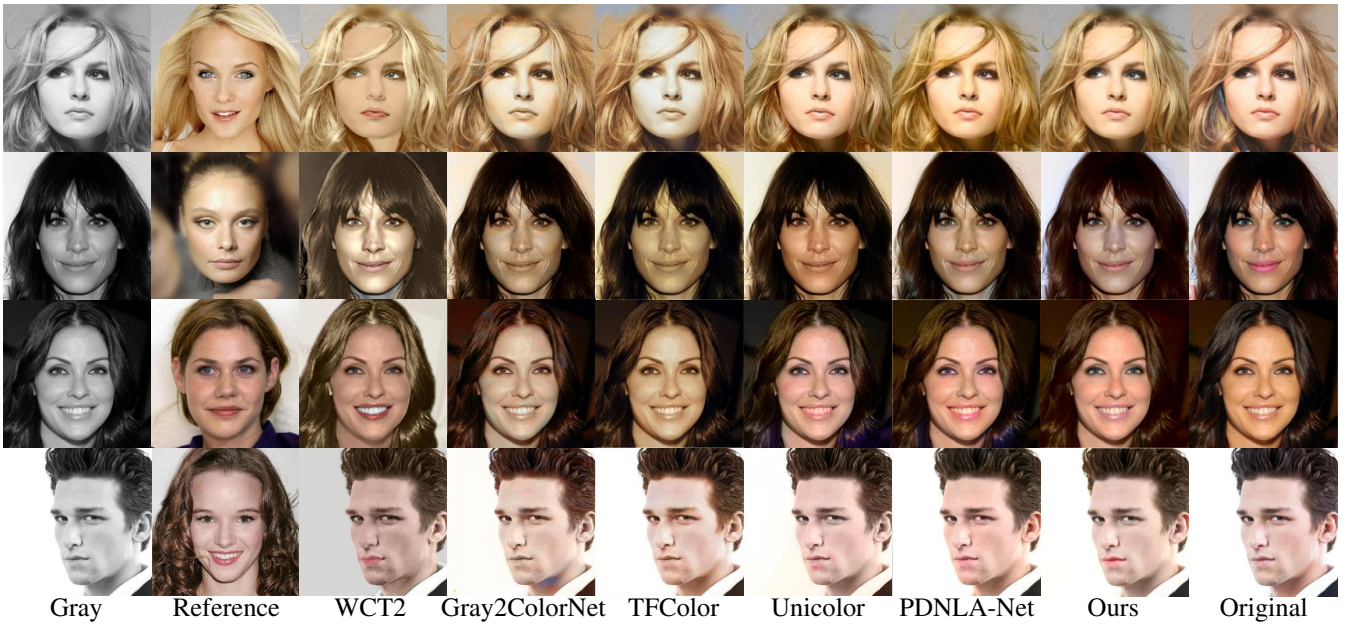

\begin{table}[t]
	\centering
	\caption{Quantitative comparison on CelebA-HQ dataset. The method ranked first for each metric is displayed in bold, and the method ranked second is underlined.}
	\label{Quantitative}
	\begin{tabular}{c|ccccc}
		\toprule
		Methods & FID↓ & CF↑ & PSNR↑ & SSIM↑  \\ \midrule 
		Colorful (2016) & 11.41  & 37.82                    & 24.69 & 0.9419 \\
		InstColor (2020) & 8.99  & 37.02                    & 25.46 & \underline{0.9940} \\
		GCP (2021)       & 10.29 & 33.70                    & 24.33 & 0.9937 \\
		Unicolor (2022) & 6.17  & 37.41                    & 23.21 & 0.9406 \\
		DisColor (2022) & 8.79  & 43.37                    & 23.72 & 0.9934 \\
		CT2 (2022) & 8.61  & 43.33                    & 24.55 & 0.9415 \\
		BigColor (2022)  & 6.34  & 42.99                    & 23.34 & 0.9339 \\
		iColoriT (2023)  & 12.68  & 32.32                    & \underline{27.53} & 0.9809 \\
		DDColor (2023)   & \underline{3.80}  &\textbf{49.71}                    & 24.25 & 0.9936 \\
		L-CAD (2024)   & 7.43  & 28.73                  & 23.86 & 0.9911 \\
		Ours      & \textbf{3.60}  & \underline{43.73}                    & \textbf{30.50} & \textbf{0.9946} \\ \bottomrule
	\end{tabular}
\end{table}
\subsection{Comparison with Previous Methods}
\paragraph{Quantitative comparison.} We evaluate the performance of the automatic colorization results of our proposed method against prior techniques on the test dataset, reporting quantitative results across four distinct metrics in ~\cref{Quantitative}. 
The proposed method is evaluated against a selection of the most recent and representative colorization techniques, including Colorful~\cite{ColorfulColorization}, InstColor~\cite{Instcolor}, GCP~\cite{GCP}, Unicolor~\cite{unicolor}, DisColor~\cite{DisentangledColorization}, CT2~\cite{ct2}, BigColor~\cite{BigColor}, iColoriT~\cite{iColoriT}, DDColor~\cite{DDcolor}, and L-CAD~\cite{LCAD}.
Experimental evaluations utilize the official codebases and pre-trained model weights provided by the respective authors, as well as the automatic colorization approaches described in their papers. The proposed method achieves the lowest FID, indicating superior colorization quality and fidelity. In terms of colorfulness score, which reflects color richness, the approach ranks second among the evaluated techniques. However, excessive colorfulness does not necessarily equate to optimal visual quality.
Our proposed method exhibits the best performance on the PSNR and SSIM pixel-wise evaluation metrics, demonstrating the high quality of the generated images.

\paragraph{Qualitative evaluation.} 
\cref{fig:Qualitative} showcases representative automatic colorization results from our baseline methods. Note that the original colored images serve as reference benchmarks rather than definitive standards. Our results demonstrate the minimal degree of color bleeding and the most naturalistic visual appearance. Furthermore, our colorization outputs do not present an excessive variety of colors that would lead to an unrealistic appearance. 
\cref{fig:Qualitative_ref} presents qualitative comparisons between the proposed method's reference image-guided colorization results and those of the latest representative reference image-guided colorization techniques, including WCT2~\cite{wct2}, Gray2ColorNet~\cite{Gray2colornet}, TFcolor~\cite{TFcolor}, Unicolor~\cite{unicolor}, and PDNLA-Net~\cite{PDNLA-Net}.
Early techniques like WCT2~\cite{wct2} fail to consider the input image's grayscale values. Global color transfer methods, exemplified by Gray2Colornet~\cite{Gray2colornet}, exhibit difficulty accurately specifying individual component hues. Unicolor~\cite{unicolor} and PDNLA-Net~\cite{PDNLA-Net} concomitantly exhibit an inability to accurately specify intricate details, such as iris hue, culminating in undesirable color bleeding. 
Furthermore, to demonstrate the visual effects of our method under various color control approaches, we have introduced more reference image-guided colorization results guided by referencing single or multiple image(s) and diverse colorization results guided by single or multiple sampling(s) from Gaussian distribution. These additional results are included in the supplementary material.

\begin{table*}[!tb]
	\centering
	\caption{Quantitative comparison on Ablation studies. The method ranked first for each metric is displayed in bold, and the method ranked second is underlined.}
	\label{Abalation}
	\begin{tabular}{cccccccccc}
		\toprule
		\multirow{3}{*}{Models} &
		\multirow{3}{*}{\begin{tabular}[c]{@{}c@{}}Color \\Representation \\Branch \end{tabular}} &
		\multicolumn{2}{c}{Data Augmentation} &
		\multirow{3}{*}{\begin{tabular}[c]{@{}c@{}}Grouped \\ design\end{tabular}} &
		\multirow{3}{*}{FID↓} &
		\multirow{3}{*}{CF↑} &
		\multirow{3}{*}{PSNR↑} &
		\multirow{3}{*}{SSIM↑} \\ \cline{3-4}
		&
		&
		\begin{tabular}[c]{@{}c@{}}Chromatic\\ Augmentation\end{tabular} &
		\begin{tabular}[c]{@{}c@{}}Spatial\\ Augmentation\end{tabular} &
		&
		&
		&
		&
		\\ \midrule
		Baseline                            &           &           &           &           & 10.43 & 43.07  & 24.53 & 0.9937 \\ \midrule
		\multirow{3}{*}{w/o Data Augmentaion} & \checkmark &           &           & \checkmark & 7.84 & 42.51 & 25.89 & 0.9942 \\
		& \checkmark & \checkmark &           & \checkmark & 7.32    &  \textbf{43.92}   &  \underline{28.01}    &  \underline{0.9945}      \\
		& \checkmark &           & \checkmark & \checkmark & 5.45    &   43.35   &  26.91    &  \textbf{0.9946}      \\ \midrule
		w/o Grouped design                 & \checkmark & \checkmark & \checkmark &           & \underline{5.38}  & 43.29  & 26.77  &  \textbf{0.9946}  \\ \midrule
		Ours                                & \checkmark & \checkmark & \checkmark & \checkmark & \textbf{3.60} & \underline{43.73} & \textbf{30.50} & \textbf{0.9946} \\ \bottomrule
	\end{tabular}
\end{table*}

\subsection{Ablation Study}
We conducted ablation studies to assess the impact of several important designs in our model and training. The quantitative results were obtained on the full test set. Unless explicitly stated, we maintained a consistent training configuration during ablation studies.

\paragraph{Color representation branch.} 
The proposed color representation branch $g$ decouples the color information from the facial components to obtain the color representation $\mat{w}$. 

Following the removal of the color representation decoder $G_{w}$ structure associated with the $g$, the remaining $E_{Gray}$ and $G$ serve as the baseline for the subsequent ablation study.
As evidenced in ~\cref{Abalation}, the quantitative results of the automatic colorization approach underscore the crucial role of the color representation branch in enhancing the various quantitative performance metrics. Please refer to the supplementary materials for the qualitative results.

\paragraph{Data Augmentation.} 
During the training phase, to obtain a more effectively decoupled color representation, we employed chromatic augmentation techniques such as color jittering, as well as spatial augmentation approaches including affine transformations and flipping. We conduct ablation studies to demonstrate the necessity of these two types of augmentation and to elucidate their respective contributions.
~\cref{Abalation} demonstrates that chromatic augmentation enhanced the CF metric, indicating its capacity to expand the network's color coverage range. Conversely, spatial augmentation reduced the FID metric, showcasing its ability to improve the fidelity of the generated images. Employing both augmentation approaches simultaneously optimized these two metrics, establishing a favorable trade-off.
Please refer to the supplementary materials for the qualitative results.

\paragraph{Grouped design.} The $G_{w}$ module is designed with a grouped structure to process the individual parts of $\mat{w}$ separately. To evaluate the contribution of this design, we conducted comparative experiments, as detailed in ~\cref{Abalation}.
We performed comparative experiments to assess the impact of this architectural design on the network's controllability, as detailed in \cref{Abalation}.
The results demonstrate that omitting the grouped design leads to deficiencies across various quantitative metrics, compared to our intact approach. Furthermore, the qualitative analysis presented in the supplementary materials reveals that irrespective of the coloring method, removing the grouped structure results in color bleeding and spills onto other regions of the face.

\section{Conclusion}
\label{sec:Conclusion}
In this work, we have developed a novel framework for component-specific facial colorization that offers advanced control and versatility. The core innovation of our approach lies in the decoupling of facial component colors into a dedicated color representation within our color representation branch. This design choice enables the independent specification of colors for each facial component, thereby facilitating diverse color control.

\bibliography{aaai25}

\begin{thebibliography}{50}
\providecommand{\natexlab}[1]{#1}

\bibitem[{Bahng et~al.(2018)Bahng, Yoo, Cho, Park, Wu, Ma, and Choo}]{bahng2018coloring}
Bahng, H.; Yoo, S.; Cho, W.; Park, D.~K.; Wu, Z.; Ma, X.; and Choo, J. 2018.
\newblock Coloring with words: Guiding image colorization through text-based palette generation.
\newblock In \emph{Proceedings of the european conference on computer vision (eccv)}, 431--447.

\bibitem[{Chang et~al.(2015)Chang, Fried, Liu, DiVerdi, and Finkelstein}]{chang2015palette}
Chang, H.; Fried, O.; Liu, Y.; DiVerdi, S.; and Finkelstein, A. 2015.
\newblock Palette-based photo recoloring.
\newblock \emph{ACM Trans. Graph.}, 34(4): 139--1.

\bibitem[{Charpiat, Hofmann, and Sch{\"o}lkopf(2008)}]{charpiat2008automatic}
Charpiat, G.; Hofmann, M.; and Sch{\"o}lkopf, B. 2008.
\newblock Automatic image colorization via multimodal predictions.
\newblock In \emph{Computer Vision--ECCV 2008: 10th European Conference on Computer Vision, Marseille, France, October 12-18, 2008, Proceedings, Part III 10}, 126--139. Springer.

\bibitem[{Cheng, Yang, and Sheng(2015)}]{cheng2015deep}
Cheng, Z.; Yang, Q.; and Sheng, B. 2015.
\newblock Deep colorization.
\newblock In \emph{Proceedings of the IEEE international conference on computer vision}, 415--423.

\bibitem[{Chia et~al.(2011)Chia, Zhuo, Gupta, Tai, Cho, Tan, and Lin}]{chia2011semantic}
Chia, A. Y.-S.; Zhuo, S.; Gupta, R.~K.; Tai, Y.-W.; Cho, S.-Y.; Tan, P.; and Lin, S. 2011.
\newblock Semantic colorization with internet images.
\newblock \emph{ACM Transactions on Graphics (ToG)}, 30(6): 1--8.

\bibitem[{Deshpande et~al.(2017)Deshpande, Lu, Yeh, Jin~Chong, and Forsyth}]{deshpande2017learning}
Deshpande, A.; Lu, J.; Yeh, M.-C.; Jin~Chong, M.; and Forsyth, D. 2017.
\newblock Learning diverse image colorization.
\newblock In \emph{Proceedings of the IEEE conference on computer vision and pattern recognition}, 6837--6845.

\bibitem[{Dinh, Krueger, and Bengio(2014)}]{NICE}
Dinh, L.; Krueger, D.; and Bengio, Y. 2014.
\newblock Nice: Non-linear independent components estimation.
\newblock \emph{arXiv preprint arXiv:1410.8516}.

\bibitem[{Dinh, Sohl-Dickstein, and Bengio(2016)}]{RealNVP}
Dinh, L.; Sohl-Dickstein, J.; and Bengio, S. 2016.
\newblock Density estimation using real nvp.
\newblock \emph{arXiv preprint arXiv:1605.08803}.

\bibitem[{Dosovitskiy et~al.(2020)Dosovitskiy, Beyer, Kolesnikov, Weissenborn, Zhai, Unterthiner, Dehghani, Minderer, Heigold, Gelly et~al.}]{dosovitskiy2020image}
Dosovitskiy, A.; Beyer, L.; Kolesnikov, A.; Weissenborn, D.; Zhai, X.; Unterthiner, T.; Dehghani, M.; Minderer, M.; Heigold, G.; Gelly, S.; et~al. 2020.
\newblock An image is worth 16x16 words: Transformers for image recognition at scale.
\newblock \emph{arXiv preprint arXiv:2010.11929}.

\bibitem[{Gupta et~al.(2012)Gupta, Chia, Rajan, Ng, and Zhiyong}]{gupta2012image}
Gupta, R.~K.; Chia, A. Y.-S.; Rajan, D.; Ng, E.~S.; and Zhiyong, H. 2012.
\newblock Image colorization using similar images.
\newblock In \emph{Proceedings of the 20th ACM international conference on Multimedia}, 369--378.

\bibitem[{He et~al.(2018)He, Chen, Liao, Sander, and Yuan}]{he2018deep}
He, M.; Chen, D.; Liao, J.; Sander, P.~V.; and Yuan, L. 2018.
\newblock Deep exemplar-based colorization.
\newblock \emph{ACM Transactions on Graphics (TOG)}, 37(4): 1--16.

\bibitem[{Heusel et~al.(2017)Heusel, Ramsauer, Unterthiner, Nessler, and Hochreiter}]{heusel2017gans}
Heusel, M.; Ramsauer, H.; Unterthiner, T.; Nessler, B.; and Hochreiter, S. 2017.
\newblock Gans trained by a two time-scale update rule converge to a local nash equilibrium.
\newblock \emph{Advances in neural information processing systems}, 30.

\bibitem[{Huang, Zhao, and Liao(2022)}]{unicolor}
Huang, Z.; Zhao, N.; and Liao, J. 2022.
\newblock UniColor: A Unified Framework for Multi-Modal Colorization with Transformer.
\newblock \emph{ACM Trans. Graph.}, 41(6).

\bibitem[{Huynh-Thu and Ghanbari(2008)}]{huynh2008scope}
Huynh-Thu, Q.; and Ghanbari, M. 2008.
\newblock Scope of validity of PSNR in image/video quality assessment.
\newblock \emph{Electronics letters}, 44(13): 800--801.

\bibitem[{Iizuka and Simo-Serra(2019)}]{iizuka2019deepremaster}
Iizuka, S.; and Simo-Serra, E. 2019.
\newblock Deepremaster: temporal source-reference attention networks for comprehensive video enhancement.
\newblock \emph{ACM Transactions on Graphics (TOG)}, 38(6): 1--13.

\bibitem[{Ironi, Cohen-Or, and Lischinski(2005)}]{ironi2005colorization}
Ironi, R.; Cohen-Or, D.; and Lischinski, D. 2005.
\newblock Colorization by Example.
\newblock \emph{Rendering techniques}, 29: 201--210.

\bibitem[{Ji et~al.(2022)Ji, Jiang, Luo, Tao, Chu, Xie, Wang, and Tai}]{ji2022colorformer}
Ji, X.; Jiang, B.; Luo, D.; Tao, G.; Chu, W.; Xie, Z.; Wang, C.; and Tai, Y. 2022.
\newblock ColorFormer: Image colorization via color memory assisted hybrid-attention transformer.
\newblock In \emph{European Conference on Computer Vision}, 20--36. Springer.

\bibitem[{Johnson, Alahi, and Fei-Fei(2016)}]{johnson2016perceptual}
Johnson, J.; Alahi, A.; and Fei-Fei, L. 2016.
\newblock Perceptual losses for real-time style transfer and super-resolution.
\newblock In \emph{Computer Vision--ECCV 2016: 14th European Conference, Amsterdam, The Netherlands, October 11-14, 2016, Proceedings, Part II 14}, 694--711. Springer.

\bibitem[{Kang et~al.(2023)Kang, Yang, Ouyang, Ren, Li, and Xie}]{DDcolor}
Kang, X.; Yang, T.; Ouyang, W.; Ren, P.; Li, L.; and Xie, X. 2023.
\newblock Ddcolor: Towards photo-realistic image colorization via dual decoders.
\newblock In \emph{Proceedings of the IEEE/CVF International Conference on Computer Vision}, 328--338.

\bibitem[{Karras et~al.(2017)Karras, Aila, Laine, and Lehtinen}]{karras2017progressive}
Karras, T.; Aila, T.; Laine, S.; and Lehtinen, J. 2017.
\newblock Progressive growing of gans for improved quality, stability, and variation.
\newblock \emph{arXiv preprint arXiv:1710.10196}.

\bibitem[{Karras, Laine, and Aila(2019)}]{karras2019style}
Karras, T.; Laine, S.; and Aila, T. 2019.
\newblock A style-based generator architecture for generative adversarial networks.
\newblock In \emph{Proceedings of the IEEE/CVF conference on computer vision and pattern recognition}, 4401--4410.

\bibitem[{Ke et~al.(2023)Ke, Liu, Zhu, Zhao, and Lau}]{ke2023neural}
Ke, Z.; Liu, Y.; Zhu, L.; Zhao, N.; and Lau, R.~W. 2023.
\newblock Neural preset for color style transfer.
\newblock In \emph{Proceedings of the IEEE/CVF Conference on Computer Vision and Pattern Recognition}, 14173--14182.

\bibitem[{Kim et~al.(2022)Kim, Kang, Kim, Lee, Kim, Kim, Baek, and Cho}]{BigColor}
Kim, G.; Kang, K.; Kim, S.; Lee, H.; Kim, S.; Kim, J.; Baek, S.-H.; and Cho, S. 2022.
\newblock BigColor: Colorization using a generative color prior for natural images.
\newblock In \emph{European Conference on Computer Vision}, 350--366. Springer.

\bibitem[{Kingma and Ba(2014)}]{kingma2014adam}
Kingma, D.~P.; and Ba, J. 2014.
\newblock Adam: A method for stochastic optimization.
\newblock \emph{arXiv preprint arXiv:1412.6980}.

\bibitem[{Kingma and Dhariwal(2018)}]{GLOW}
Kingma, D.~P.; and Dhariwal, P. 2018.
\newblock Glow: Generative flow with invertible 1x1 convolutions.
\newblock \emph{Advances in neural information processing systems}, 31.

\bibitem[{Kumar, Weissenborn, and Kalchbrenner(2021)}]{kumar2021colorization}
Kumar, M.; Weissenborn, D.; and Kalchbrenner, N. 2021.
\newblock Colorization transformer.
\newblock \emph{arXiv preprint arXiv:2102.04432}.

\bibitem[{Levin, Lischinski, and Weiss(2004)}]{levin2004colorization}
Levin, A.; Lischinski, D.; and Weiss, Y. 2004.
\newblock Colorization using optimization.
\newblock In \emph{ACM SIGGRAPH 2004 Papers}, 689--694.

\bibitem[{Liang et~al.(2024)Liang, Li, Zhou, Li, and Loy}]{liang2024control}
Liang, Z.; Li, Z.; Zhou, S.; Li, C.; and Loy, C.~C. 2024.
\newblock Control Color: Multimodal Diffusion-based Interactive Image Colorization.
\newblock \emph{arXiv preprint arXiv:2402.10855}.

\bibitem[{Liu et~al.(2008)Liu, Wan, Qu, Wong, Lin, Leung, and Heng}]{liu2008intrinsic}
Liu, X.; Wan, L.; Qu, Y.; Wong, T.-T.; Lin, S.; Leung, C.-S.; and Heng, P.-A. 2008.
\newblock Intrinsic colorization.
\newblock In \emph{ACM SIGGRAPH Asia 2008 papers}, 1--9.

\bibitem[{Lu et~al.(2020)Lu, Yu, Peng, Zhao, and Wang}]{Gray2colornet}
Lu, P.; Yu, J.; Peng, X.; Zhao, Z.; and Wang, X. 2020.
\newblock Gray2colornet: Transfer more colors from reference image.
\newblock In \emph{Proceedings of the 28th ACM international conference on multimedia}, 3210--3218.

\bibitem[{Qu, Wong, and Heng(2006)}]{qu2006manga}
Qu, Y.; Wong, T.-T.; and Heng, P.-A. 2006.
\newblock Manga colorization.
\newblock \emph{ACM Transactions on Graphics (ToG)}, 25(3): 1214--1220.

\bibitem[{Su, Chu, and Huang(2020)}]{Instcolor}
Su, J.-W.; Chu, H.-K.; and Huang, J.-B. 2020.
\newblock Instance-aware image colorization.
\newblock In \emph{Proceedings of the IEEE/CVF Conference on Computer Vision and Pattern Recognition}, 7968--7977.

\bibitem[{Tsaftaris et~al.(2014)Tsaftaris, Casadio, Andral, and Katsaggelos}]{tsaftaris2014novel}
Tsaftaris, S.~A.; Casadio, F.; Andral, J.-L.; and Katsaggelos, A.~K. 2014.
\newblock A novel visualization tool for art history and conservation: Automated colorization of black and white archival photographs of works of art.
\newblock \emph{Studies in conservation}, 59(3): 125--135.

\bibitem[{Vitoria, Raad, and Ballester(2020)}]{vitoria2020chromagan}
Vitoria, P.; Raad, L.; and Ballester, C. 2020.
\newblock Chromagan: Adversarial picture colorization with semantic class distribution.
\newblock In \emph{Proceedings of the IEEE/CVF winter conference on applications of computer vision}, 2445--2454.

\bibitem[{Wang et~al.(2023)Wang, Zhai, Liu, Jiang, and Gao}]{PDNLA-Net}
Wang, H.; Zhai, D.; Liu, X.; Jiang, J.; and Gao, W. 2023.
\newblock Unsupervised deep exemplar colorization via pyramid dual non-local attention.
\newblock \emph{IEEE Transactions on Image Processing}.

\bibitem[{Wang et~al.(2004)Wang, Bovik, Sheikh, and Simoncelli}]{wang2004image}
Wang, Z.; Bovik, A.~C.; Sheikh, H.~R.; and Simoncelli, E.~P. 2004.
\newblock Image quality assessment: from error visibility to structural similarity.
\newblock \emph{IEEE transactions on image processing}, 13(4): 600--612.

\bibitem[{Weng et~al.(2022)Weng, Sun, Li, Li, and Shi}]{ct2}
Weng, S.; Sun, J.; Li, Y.; Li, S.; and Shi, B. 2022.
\newblock CT 2: Colorization transformer via color tokens.
\newblock In \emph{European Conference on Computer Vision}, 1--16. Springer.

\bibitem[{Weng et~al.(2024)Weng, Zhang, Li, Li, Shi et~al.}]{LCAD}
Weng, S.; Zhang, P.; Li, Y.; Li, S.; Shi, B.; et~al. 2024.
\newblock L-cad: Language-based colorization with any-level descriptions using diffusion priors.
\newblock \emph{Advances in Neural Information Processing Systems}, 36.

\bibitem[{Wu et~al.(2021)Wu, Wang, Li, Zhang, Zhao, and Shan}]{GCP}
Wu, Y.; Wang, X.; Li, Y.; Zhang, H.; Zhao, X.; and Shan, Y. 2021.
\newblock Towards vivid and diverse image colorization with generative color prior.
\newblock In \emph{Proceedings of the IEEE/CVF international conference on computer vision}, 14377--14386.

\bibitem[{Xia et~al.(2022)Xia, Hu, Wong, and Wang}]{DisentangledColorization}
Xia, M.; Hu, W.; Wong, T.-T.; and Wang, J. 2022.
\newblock Disentangled image colorization via global anchors.
\newblock \emph{ACM Transactions on Graphics (TOG)}, 41(6): 1--13.

\bibitem[{Xu et~al.(2020)Xu, Wang, Fang, Sheng, and Zhang}]{xu2020stylization}
Xu, Z.; Wang, T.; Fang, F.; Sheng, Y.; and Zhang, G. 2020.
\newblock Stylization-based architecture for fast deep exemplar colorization.
\newblock In \emph{Proceedings of the IEEE/CVF Conference on Computer Vision and Pattern Recognition}, 9363--9372.

\bibitem[{Yin et~al.(2021)Yin, Lu, Zhao, and Peng}]{TFcolor}
Yin, W.; Lu, P.; Zhao, Z.; and Peng, X. 2021.
\newblock Yes," Attention Is All You Need", for Exemplar based Colorization.
\newblock In \emph{Proceedings of the 29th ACM international conference on multimedia}, 2243--2251.

\bibitem[{Yoo et~al.(2019)Yoo, Uh, Chun, Kang, and Ha}]{wct2}
Yoo, J.; Uh, Y.; Chun, S.; Kang, B.; and Ha, J.-W. 2019.
\newblock Photorealistic style transfer via wavelet transforms.
\newblock In \emph{Proceedings of the IEEE/CVF international conference on computer vision}, 9036--9045.

\bibitem[{Yun et~al.(2023)Yun, Lee, Park, and Choo}]{iColoriT}
Yun, J.; Lee, S.; Park, M.; and Choo, J. 2023.
\newblock iColoriT: Towards propagating local hints to the right region in interactive colorization by leveraging vision transformer.
\newblock In \emph{Proceedings of the IEEE/CVF Winter Conference on Applications of Computer Vision}, 1787--1796.

\bibitem[{Zhang, Isola, and Efros(2016)}]{ColorfulColorization}
Zhang, R.; Isola, P.; and Efros, A.~A. 2016.
\newblock Colorful image colorization.
\newblock In \emph{Computer Vision--ECCV 2016: 14th European Conference, Amsterdam, The Netherlands, October 11-14, 2016, Proceedings, Part III 14}, 649--666. Springer.

\bibitem[{Zhang et~al.(2017)Zhang, Zhu, Isola, Geng, Lin, Yu, and Efros}]{zhang2017real}
Zhang, R.; Zhu, J.-Y.; Isola, P.; Geng, X.; Lin, A.~S.; Yu, T.; and Efros, A.~A. 2017.
\newblock Real-time user-guided image colorization with learned deep priors.
\newblock \emph{arXiv preprint arXiv:1705.02999}.

\bibitem[{Zhang et~al.(2023)Zhang, Wei, Jiang, Zhang, Zuo, and Tian}]{zhang2023controlvideo}
Zhang, Y.; Wei, Y.; Jiang, D.; Zhang, X.; Zuo, W.; and Tian, Q. 2023.
\newblock Controlvideo: Training-free controllable text-to-video generation.
\newblock \emph{arXiv preprint arXiv:2305.13077}.

\bibitem[{Zhao et~al.(2020)Zhao, Han, Shao, and Snoek}]{zhao2020pixelated}
Zhao, J.; Han, J.; Shao, L.; and Snoek, C.~G. 2020.
\newblock Pixelated semantic colorization.
\newblock \emph{International Journal of Computer Vision}, 128: 818--834.

\bibitem[{Zhao et~al.(2018)Zhao, Liu, Snoek, Han, and Shao}]{zhao2018pixel}
Zhao, J.; Liu, L.; Snoek, C.~G.; Han, J.; and Shao, L. 2018.
\newblock Pixel-level semantics guided image colorization.
\newblock \emph{arXiv preprint arXiv:1808.01597}.

\bibitem[{Zhu et~al.(2017)Zhu, Park, Isola, and Efros}]{zhu2017unpaired}
Zhu, J.-Y.; Park, T.; Isola, P.; and Efros, A.~A. 2017.
\newblock Unpaired image-to-image translation using cycle-consistent adversarial networks.
\newblock In \emph{Proceedings of the IEEE international conference on computer vision}, 2223--2232.

\end{thebibliography}

\end{document}